\documentclass[11pt,a4paper]{article}
\usepackage[hyperref]{naaclhlt2019}
\usepackage{times}
\usepackage{latexsym}
\usepackage{amssymb,amsmath}
\usepackage{url}
\usepackage{caption}
\usepackage{graphicx}

\usepackage{graphics}

\aclfinalcopy % Uncomment this line for the final submission
\title{Identification, Interpretability, and Bayesian Word Embeddings}

\author{Adam M. Lauretig \\
Department of Political Science \\
  Ohio State University  \\
  Columbus, OH \\
  {\tt lauretig.1@osu.edu}  \\}

\date{}

\begin{document}
\maketitle
\begin{abstract}

Social scientists have recently turned to analyzing text using tools from natural language processing like word embeddings to measure concepts like ideology, bias, and affinity. However, word embeddings are difficult to use in the regression framework familiar to social scientists: embeddings are are neither identified, nor directly interpretable. I offer two advances on standard embedding models to remedy these problems. First, I develop Bayesian Word Embeddings with Automatic Relevance Determination priors, relaxing the assumption that all embedding dimensions have equal weight. Second, I apply work identifying latent variable models to anchor the dimensions of the resulting embeddings, identifying them, and making them interpretable and usable in a regression. I then apply this model and anchoring approach to two cases, the shift in internationalist rhetoric in the American presidents' inaugural addresses, and the relationship between bellicosity in American foreign policy decision-makers' deliberations. I find that inaugural addresses became less internationalist after 1945, which goes against the conventional wisdom, and that an increase in bellicosity is associated with an increase in hostile actions by the United States, showing that elite deliberations are not cheap talk, and helping confirm the validity of the model.

\end{abstract}

\section{Introduction}

Important questions in the social sciences turn on the meanings of words used to express ideas like language change, emotion, and ideological affinity \citep{hamilton2016diachronic, rheault2016measuring, pomeroy2018multiplex}. One increasingly popular way to represent meaning, originating in natural language processing, is through the use of word embeddings. This class of models learns a set of coefficients which encode meaning by predicting a word given the surrounding words \citep{mikolov2013distributed, mikolov2013linguistic}. These coefficients are the \emph{embeddings}, which can then be used to analyze word meanings. 

Unfortunately, existing embedding models are not always appropriate for answering social scientists' questions. Embeddings are not identified, and the dimensions are not directly interpretable, which makes it difficult to perform statistical inference on the embeddings produced by standard models, for example, using them as covariates in a regression model.\footnote{This is because in a regression setup, the coefficient is the change in the dependent variable for a 1-unit increase in the independent variable. With embedding dimensions, it is not clear what a 1-unit increase in the independent variable means, nor does direction have any clear meaning.}

To resolve these issues, I cast word embeddings as a Bayesian latent variable model. Identifying multidimensional latent variable models is a known problem, and I draw on solutions proposed in the ideal point modeling literature \citep{rivers2003identification, clinton2004statistical} to render embeddings interpretable and usable in a regression framework. I demonstrate these results on two corpora: a collection of inaugural addresses, and a selection of declassified diplomatic documents from the \emph{Foreign Relation of the United States}. In the inaugural addresses, I find rhetoric became more domestically-focused after 1945, a shift which existing social science approaches cannot detect. This finding stands in contrast to what existing theories of international relations would have us expect. In the \textit{FRUS} documents, I find that more bellicose rhetoric results in more aggressive American foreign policy behavior, helping confirm that elite deliberation matters for shaping foreign policy, and that the measurements I create correlate with existing datasets, helping to establish the validity of the model results.

\section{Social Science and Embedding Models of Language}

Traditional approaches to creating variables from text in the social sciences involve human coders, who assign documents to categories based on pre-defined criteria. However, this approach is expensive, and does not scale. Text as data techniques attempt to solve this problem through the use of natural language processing techniques to convert a corpus of text into numeric objects which makes inference possible \citep{grimmer2013text, gentzkow2017text}. These techniques allow scholars to create variables and operationalize concepts in corpora that are too large for human coding, and investigate ideas which cannot be measured directly (unlike indicators like Gross Domestic Product or population).

While a variety of models have been proposed to create variables from political text, including scaling models \citep{lowe2011scaling}, and topic models \citep{blei2003latent, grimmer2010bayesian, roberts2016model}, these approaches focus on the document as the unit of analysis. Word embeddings, which have a long history in the natural language processing literature (see \citet{turney2010frequency} for an extensive review of pre-neural network models), have recently been embraced by social scientists for their potential for inference at the word level. Modern neural word embedding models learn a low-dimensional representation of a word as a dense vector by either factorizing a word co-occurrence matrix or predicting the co-occurrence of a pair of words using a single-layer neural network. Among the best known of these models is word2vec \citep{mikolov2013distributed, mikolov2013linguistic}, which proposed an efficient model for learning embeddings, framing embedding learning as a prediction task, rather than a factorization task. 

For social scientists, word embeddings are a powerful tool because they can represent the meanings of individual words. Embeddings can help isolate patterns in corpora that are expensive to label, and make apparent latent phenomena not observable through simple document-feature counts such as patterns of semantic change, \citep{hamilton2016diachronic}, cultural assumptions and biases \citep{caliskan2017semantics, kozlowski2018geometry, garg2018word}, and ideological affinity in international organizations \citep{pomeroy2018multiplex}.

However, these embeddings can be problematic for social science research, where scholars care about both model identification and interpretable results. Embeddings are multidimensional latent variable models, which are not, by default, identified: a known problem with this class of model, where multiple permutations of latent dimensions can result in the same observed data \citep{rivers2003identification, clinton2004statistical, aldrich2014polarization}. However, by anchoring points on these dimensions, it is possible to present identified and interpretable dimensions. In the ideal-point literature, these anchors represent ideological ``endpoints,'' with theory guiding the selection of which legislators are most liberal and conservative. Choosing words as anchors with a large number of dimensions is more difficult than choosing legislators, however, I offer a solution below.

There have been multiple efforts at developing Bayesian word embeddings \citep{rudolph2016exponential, barkan2017bayesian, ji2017bayesian, havrylov2018embedding}, however, none of these have exploited the key advantage of Bayesian inference: the ability to quantify the uncertainty in parameter estimates, and use prior information to inform parameter estimates. The one approach that has incorporated both uncertainty and hypothesis testing is \citet{han2018conditional}, who offer both measures of uncertainty, and a way to test the effect of metadata on the similarity of embeddings, however, this approach does not account for identification problems in the learned embeddings. 
%Works which have examined the effects of language on an outcome, and provide interpretable results have focused on corpus level topics, rather than individual words \citep{fong2016discovery, egami2018make, mozer2018matching}.

\section{Bayesian Word Embeddings}

In this section, I develop word embeddings as Bayesian latent variable models estimated with variational inference, following similar work for probabilistic principal components analysis \citep{bishop1999variational} and ideal-point models \citep{imai2016fast}. I first discuss the embedding model setup, add Automatic Relevance Determination priors to the model, and then, present the variational updates to estimate the model. 

Word embeddings predict the probability of a word-context pair co-occurring, and because the co-occurrence is a binary variable ($Y_{ij} = 1$ if $w_{i}$ and $w_{j}$ co-occur, $0$ otherwise), I use a probit link function to model the probability of co-occurrence.
\begingroup\makeatletter\def\f@size{10}\check@mathfonts
\begin{equation} 
\begin{split}
p(&Y_{ij} = 1) = \\ 
& (\pmb{1}[z_{ij} > 0]\pmb{1}[y_{ij} = 1] + 
 \pmb{1}[z_{ij} <  0]\pmb{1}[y_{ij} = 0]) \\ &\mathcal{TN}(z_{ij}|\pmb{x}_{i}^{\top}\pmb{\beta}_{j}, 1).
\end{split}
 \end{equation} 
 \endgroup
 \noindent$\pmb{X}$ and $\pmb{\beta}$ are $K \times I$ (or $K \times J$, respectively) -dimensional matrices, $\pmb{Y}$ is an $I \times J$ co-occurrence matrix, the corpus contains $I$ words and $J$ context words. Each embedding vector ($\pmb{x}_{i} \text{ or } \pmb{\beta}_{j} $) has a $K$-dimensional multivariate normal prior.

Most existing approaches to word embeddings contain no measures of uncertainty, or the covariance between dimensions. This can be a problem during estimation, as the model attempts to put equal weight on all dimensions. To resolve this, I use Automatic Relevance Determination (ARD) priors, which place a separate gamma-distributed scalar (e.g. $ \alpha_{X_{k}}$) on the diagonal for each dimension of the covariance matrix \citep{mackay1994automatic, bishop1999variational}. These priors penalize unnecessary model dimensions, improving model fit. 

This specification results in the following likelihood:
\begingroup\makeatletter\def\f@size{10}\check@mathfonts
\begin{equation}
\begin{split}
  p(\mathbf{\pmb{Z}}, & \mathbf{ \pmb{X},  \pmb{\beta}, \pmb{\alpha}_{X},  \pmb{\alpha}_{\beta}  | Y}) \propto \\
& (\pmb{1}[z_{ij} > 0]\pmb{1}[y_{ij} = 1] + 
 \pmb{1}[z_{ij} <  0]\pmb{1}[y_{ij} = 0]) \\ 
 &\mathcal{TN}(z_{ij}|\pmb{x}_{i}^{\top}\pmb{\beta}_{j}, 1) \times \\
 &\prod_i \mathcal{MVN}(\pmb{x}_i | 0, \pmb{\alpha}_{X}^{-1}) \times \\
 &\prod_i \mathcal{MVN}(\pmb{\beta}_j | 0, \pmb{\alpha}_{\beta}^{-1}) \times \\
&\prod_k \text{Gam}( \alpha_{X_{k}}| c_{X_{0}}, d_{X_{0}}) \times \\
&\prod_k \text{Gam}( \alpha_{\beta_{k}}| c_{\beta_{0}}, d_{\beta_{0}}).
\end{split}
\end{equation}
\endgroup
\noindent For Bayesian models like this, the goal is to estimate posterior distributions of the parameters most likely to have produced the observed data. Given the joint density (probability of data and parameters), we want to calculate the conditional density of the parameters by evaluating the following integral (notation follows \citet{bishop1999variational}):

\begingroup\makeatletter\def\f@size{10}\check@mathfonts
\begin{equation}
P( \pmb{Y} ) = \int p(\pmb{Y}, \pmb{ \theta }) d \pmb{ \theta}
\end{equation}
\endgroup

\noindent where $\pmb{ \theta} = \{ \pmb{Z}, \pmb{X}, \pmb{\beta}, \pmb{\alpha}_{X}, \pmb{\alpha}_{\beta} \}$. This integral is analytically intractable, so we transform the integral using Jensen's inequality:
\begingroup\makeatletter\def\f@size{10}\check@mathfonts
\begin{equation}
\begin{split}
\text{ln} P(\pmb{Y}) & = \text{ln}  \int p(\pmb{Y}, \pmb{ \theta }) d \pmb{ \theta} \\
& = \text{ln} \int \mathcal{Q}( \pmb{\theta} ) \frac{P(\pmb{Y},  \pmb{ \theta} )}{\mathcal{Q}(\pmb{\theta})}  d \pmb{ \theta} \\
& \geq \int \mathcal{Q}(\pmb{\theta}) \text{ln} \frac{P(\pmb{Y},  \pmb{ \theta} )}{\mathcal{Q}(\pmb{\theta})}    d \pmb{ \theta} \\
& = \mathcal{L}( Q)
\end{split}
\end{equation}
\endgroup

\noindent where $\mathcal{L}( Q)$ is evidence lower bound (ELBO). 

The difference between the true model $P(Y)$ and variational approximation $\mathcal{L}( Q)$ can be represented is the Kullbeck-Leibler divergence:
\begingroup\makeatletter\def\f@size{10}\check@mathfonts
\begin{equation}
KL(Q||P) = - \int \mathcal{Q}(\pmb{\theta}) \text{ln} \frac{P(Y | \pmb{ \theta} )}{\mathcal{Q}(\pmb{\theta})}    d \pmb{ \theta}
\end{equation}
\endgroup
so we turn to a mean-field variational approximation to estimate the model, minimizing the Kullbeck-Leibler divergence \citep{wainwright2008graphical, blei2017variational}. This requires assuming that the approximation to the posterior can be factorized:
\begingroup\makeatletter\def\f@size{10}\check@mathfonts
\begin{equation}
\begin{split}
\mathcal{Q}( & \pmb{Z}, \pmb{X}, \pmb{\beta}, \pmb{\alpha}_{X}, \pmb{\alpha}_{\beta} ) = \\  &\mathcal{Q}(\pmb{Z}), \mathcal{Q}(\pmb{X}), \mathcal{Q}(\pmb{\beta}), \mathcal{Q}(\pmb{\alpha}_{X}), \mathcal{Q}(\pmb{\alpha}_{\beta})
\end{split}
\end{equation}
\endgroup
and that appropriate approximating distributions can be found. In this case, the requirement is met: $z_{ij}$ is approximated with a truncated normal, $\pmb{x}_{i}$ and $\pmb{\beta}_{j}$ are approximated with multivariate normals, and $\alpha_{X_{k}}$ and $\alpha_{\beta_{k}}$ are approximated with gamma distributions. This factorization and approximation can be further factorized into the following parameter updates:
\begingroup\makeatletter\def\f@size{10}\check@mathfonts
\begin{equation}
\begin{split}
z_{ij}^{*} &= \mathbb{E}[\pmb{x}_{i}^{\top}] \mathbb{E}[\pmb{\beta}_{j}] \\ 
\mathbb{E}[q(z_{ij})] &= 
\begin{cases}
  z_{ij}^{*} + \frac{\phi(z_{ij}^{*})}{\Phi(z_{ij}^{*})}      & \quad \text{if } y_{ij} = 1 \\
  z_{ij}^{*} - \frac{\phi(z_{ij}^{*})}{1 - \Phi(z_{ij}^{*})}      & \quad \text{if } y_{ij} = 0 \\
  \end{cases}\\
\pmb{A}  &= \left( \text{diag}( \mathbb{E}[ \pmb{\alpha}_{X} ])^{-1} +  \sum_{j}\mathbb{E}[\pmb{\beta}_{j}\pmb{\beta}_{j}^{\top}] \right)  \\
\pmb{a}_{i} &= \sum_{j}  \mathbb{E}[\pmb{\beta}_{j}] \mathbb{E}[z_{ij}] \\
\mathbb{E}[q(\pmb{x}_{i})] &= \pmb{A}^{-1}\pmb{a}_{i} \\
\pmb{B} &= \left( \text{diag}(  \mathbb{E}[\pmb{\alpha}_{\beta}])^{-1}+ \sum_{i}\mathbb{E}[\pmb{x}_{i} \pmb{x}_{i}^{\top}] \right) \\
\pmb{b}_{j} &= \sum_{i}  \mathbb{E}[\pmb{x}_{i}] \mathbb{E}[z_{ij}] \\
\mathbb{E}[q(\pmb{\beta}_{j}]) &= \pmb{B}^{-1} \pmb{b}_{j} \\
c_{x} &= c_{x_0} + \frac{I}{2} \\
d_{x_{k}} &= d_{x_0} + \frac{||\mathbb{E}[\pmb{x}_k]||^{2}}{2} \\
\mathbb{E}[q( \alpha_{X_{k}} )] &= \frac{c_{x}}{d_{x_{k}}} \\
c_{\beta} &= c_{\beta_0} + \frac{J}{2} \\
d_{\beta_{k}} &= d_{\beta_0} + \frac{||\mathbb{E}[\pmb{\beta}_k]||^{2}}{2} \\
\mathbb{E}[ q(\alpha_{\beta_{k}} )] &= \frac{c_{\beta}}{d_{\beta_{k}}} \\
\end{split}
\end{equation}
\endgroup
\noindent where $c_{x_0}, d_{x_0}, c_{\beta_0}, d_{\beta_0}$ are hyperparameters set by the user. Convergence is monitored via change in the ELBO, and when change drops below a user-specified threshold, the model is considered converged. This model is implemented in the R package \texttt{bwe}.\footnote{\url{https://github.com/adamlauretig/bwe}.}
 
\section{Identifying Model Output}
 The output from multidimensional latent variable models is not identified, as many possible permutations of latent values can produce the same observed data \citep{rivers2003identification}. However, by fixing $K(K+1)$ linearly independent values (anchors), users can guarantee the embedding matrix is identified \citep{rivers2003identification,clinton2004statistical,bafumi2005practical}. To determine these anchors in the ideal point modeling literature, theory drives the endpoint selection: \citet{clinton2004statistical} fix both points for Jesse Helms, Ted Kennedy, and Lincoln Chaffee as right, left, and center anchors, respectively, in a $K=2$ model. 
 
While theory should always motivate modeling choices, determining theoretically motivated anchors when $K$ ranges from 50 to 300 can be difficult. I propose a solution: theory can motivate initial anchor selection, and then, for each additional anchor, the most cosine dis-similar word is chosen as the opposite anchor. This allows the analyst to specify theoretically motivated opposites as initial anchors, and then, resulting anchors are chosen from remaining words. I provide an implementation of this algorithm in the R package \texttt{bwe}.

\section{Interpreting Model Output}

Anchoring the embeddings ensures they are identified, however, they are still not in a format which allows for ready interpretability in the regression-based models social scientists are most familiar with. To transform embeddings so that they can be used in regression, I opt for a modification of the anchoring approach discussed above. For this approach, the user specifies a pair of endpoints for a dimension, where the endpoints of interest are set to $1$ and $-1$. This can be applied to as many dimensions as necessary, and then the automatic, cosine-based anchoring is used for the rest of the dimensions. An affine transformation is then used to transform the embedding matrix relative to the chosen anchors.  

A key advantage of this approach is that because two anchors are supplied, words are scaled on this dimension. For example, while simply choosing ``war'' as an anchor results in the results words scaled according to their similarity with ``war'', setting ``war'' and ``peace'' as opposite anchors ($1$ and $-1$, respectively) allows for a measure of bellicosity in a corpus. 

This method can be applied to as many words/concepts as the user is interested in (as the automated cosine similarity will handle the other dimensions), and, of note for social scientists, each of these word scalings, which are $I \times 1$, can be multiplied by a $D \times I$ document-term matrix: $D \times I * I \times 1$, scaling the documents in a corpus according to dimensions of interest. These document values can then we used in a regression, and the coefficients can be interpreted in a straightforward way. 

%There is a caveat: this anchoring approach cannot correct underlying biases and patterns in the training data \citep{caliskan2017semantics,garg2018word}, and, as such, computational social scientists should carefully consider the nature of their training data.

\section{Inaugurals and Internationalism}

In an initial test of this model, I investigate whether the United States saw itself in a new, global role after 1945, as perceived in presidents' inaugural addresses. After 1945, the United States was the global hegemon, and international relations theory argues that this resulted in a shift in American attitudes towards the world  \citep{mearsheimer2001tragedy}. It has been shown that the public takes elite cues on various issues \citep{zaller1992nature, druckman2015governs}, and since foreign-policy is generally viewed as an elite-led phenomenon \citep{aldrich2006foreign}, I explore whether, after the second World War, inaugural addresses were more internationally focused than those before the war. 

\begin{figure}
\centering
\includegraphics[scale=0.43]{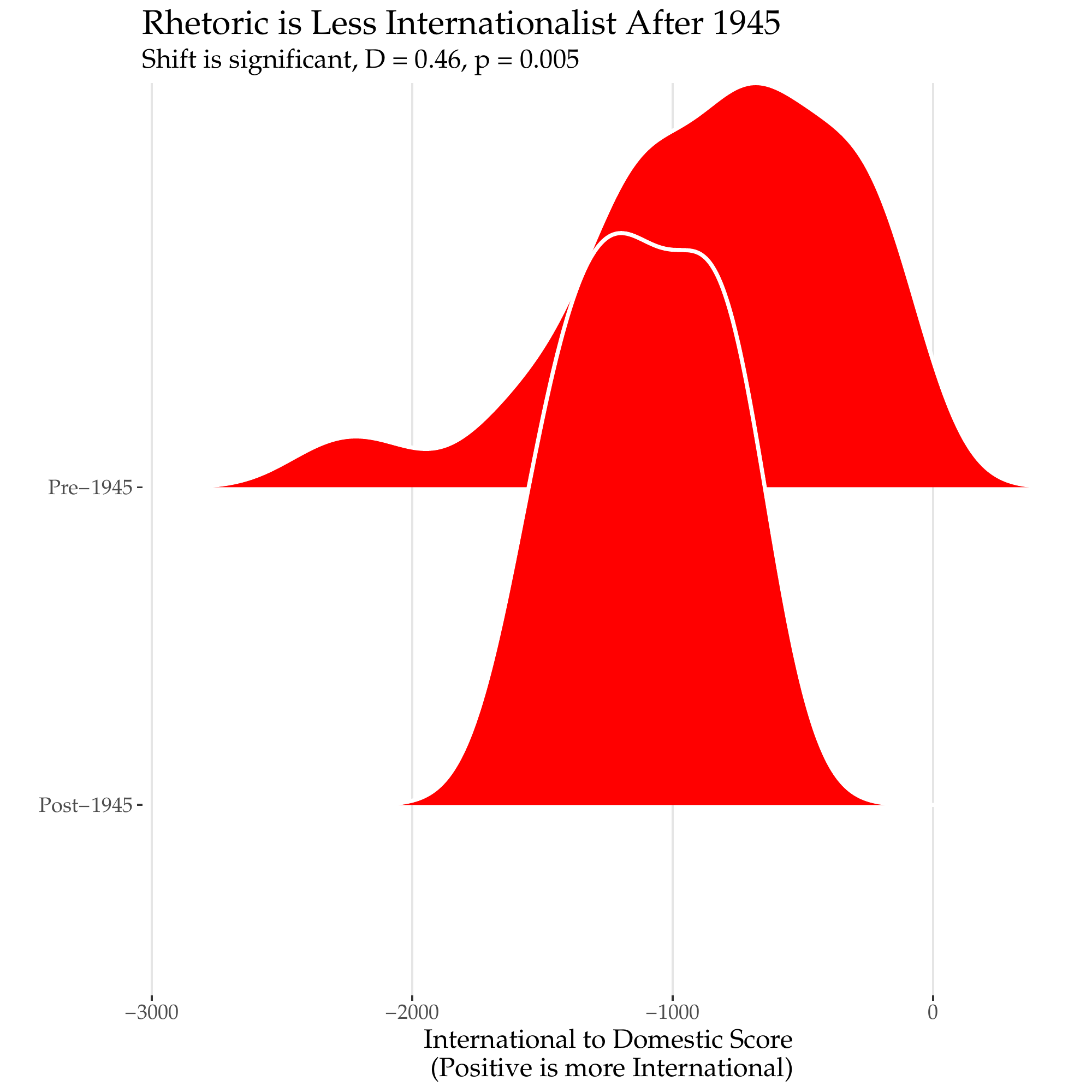}
\caption{After 1945, rhetoric in inaugural addresses becomes less internationalist, and more domestic.}
 \label{fig:inaug_diff}
\end{figure}

I use the corpus of inaugural addresses available in the \texttt{quanteda} R package \citep{benoit2018quanteda}, which contains 58 speeches. I keep words which occur with frequency $>5$, and then lowercase and tokenize the texts, resulting in 2705 words. I estimate the model with a context window of 9, with 5 negative samples for every positive sample, and and the number of dimensions $K = 50$. After fitting the model, I compare three possible anchorings: an un-anchored embedding, an embedding anchored on ``american,'' and an embedding scaled with the first dimension anchored on ``international'' and ``domestic;'' the results are visible in Table \ref{tab:inaug_sim}. We see that changing the anchoring points changes the most similar words, however, anchoring helps make these embeddings more interpretable. To test whether there was a statistically significant difference between American perceptions of global roles before and after 1945, I multiply the document-term matrix by the embedding dimension anchored on ``international'' and ``domestic,'' creating an ``internationalism'' scale for documents. I test this hypothesis using a one-sided Kolmogorov-Smirnov test, and reject the null hypothesis, that pre-1945 inaugural addresses are less internationalist than the post-1945 addresses at $p< .05$. This means pre-1945 addresses are more ``internationalist'' than the post-1945 addresses. I plot the differing distributions in Figure \ref{fig:inaug_diff}. 

What explains this finding? One possibility, building on \citet[ch. 1]{herring2008colony}, is that the United States was not isolationist prior to 1945, that isolationism was largely a product of the 1920s and 1930s, however, the United States was more unilateral before 1945. Because the multilateral world order was a fact of life after 1945, it is possible presidents were less likely to comment on international affairs, international action was the norm, rather than the exception. Furthermore, the public played a larger role in shaping foreign policy action, particularly during the Vietnam War era, than it had previously \citep{aldrich2006foreign}, and this could lead to a blurring of the lines between foreign and domestic politics when presidents address the public.

I compare the results from Bayesian Word Embeddings to the results from a standard model used in the social sciences to analyze text, the structural topic model \citep{roberts2016model}\footnote{Results presented in the Appendix.}. I find that ``domestic'' and ``international'' topics are not linked, the structural topic model captures no relationship between these words. I then investigate the change in ``domestic'' and ``international'' topics before and after 1945, and find no effect. There is ample belief and qualitative evidence of a change in American views about the world after 1945, which is not captured in the structural topic model. These differing results suggest that the embeddings are capable of recovering patterns in language that document-based topic models cannot.

\begin{table*}[t!]
\centering
\scalebox{0.7}{%
\begin{tabular}{l|llllllll}
   \hline Word of Interest  & \multicolumn{2}{l}{Anchor: International, Domestic}\\ \hline
war & large & declarations & pay & carefully & choice & equal & this & greater \\ 
  peace & practices & meeting & strife & inspiring & confederacy & advance & temple & objections \\ 
  american & engagements & soil & cultivate & by & heroes & goes & pride & she \\ 
  international & declare & path & honor & expression & speaking & where & vision & ignorance \\ 
  national & temple & subject & learned & demand & advance & objections & principle & guard \\ 
   \hline & \multicolumn{2}{l}{Anchor: American}\\ \hline
war & abroad & remedies & violate & slaves & violence & declarations & proposition & sectional \\ 
  peace & army & plenty & victory & effort & resumption & front & regulation & agreement \\ 
  american & regards & brief & instrumentality & execute & able & friendly & hands & friendship \\ 
  international & assembly & european & continent & capable & various & canal & differing & affected \\ 
  national & now & recognition & corporations & monetary & south & more & character & diversity \\ 
   \hline & \multicolumn{2}{l}{Anchor: None}\\ \hline
war & made & had & peace & force & never & after & still & place \\ 
  peace & world & nations & war & strength & prosperity & progress & just & security \\ 
  american & through & opportunity & america & life & justice & right & individual & equal \\ 
  international & maintain & lasting & fixed & beneficial & settlement & likely & relationship & intercourse \\ 
  national & most & necessity & common & given & free & first & an & power \\ 
  \end{tabular}}
\caption{The most similar words to ``war,'' ``peace,'' ``american,'' ``international,'' and ``national,'' according to each of the anchoring choices, measured via cosine similarity.
Choosing appropriate anchors leads to more interpretable embeddings than the unanchored model.  }
 \label{tab:inaug_sim}
\end{table*}

\section{Diplomacy and the Onset of War}

Natural language processing and text as data methods offer the opportunity to quantify decision-making and attitude among elites, which is notoriously difficult to measure, especially in times of conflict. Existing approaches to measuring elite attitudes often depend on survey or laboratory experiments \citep{feaver2011choosing,leveck2014role}, however, I offer an alternative approach that allows us to examine elite decisions as they occur. Drawing on a novel corpus of recently digitized diplomatic cables, the \textit{Foreign Relations of the United States} (FRUS), I investigate whether changes in the bellicosity of elite rhetoric precedes an escalation in US hostility. The FRUS dataset provides an exciting opportunity to investigate bellicosity among American foreign policy elites as events happened, as it contains primary source documents of private communications from the policymakers who develop and implement the United States' foreign policy. Among the sources for documents included in FRUS are ``Presidential libraries, Departments of State and Defense, National Security Council, Central Intelligence Agency, Agency for International Development, and other foreign affairs agencies as well as the private papers of individuals involved in formulating U.S. foreign policy,'' with a focus on documents relevant to policy-making \citep{about2017frus}. When a FRUS volume is compiled, the compiler(s) first identify a set of themes, develop a list all relevant documents, and then select those with the greatest historical import. These are then redacted or declassified, typeset, compared to the original document, and printed and bound \citep{mcallister2015toward}. 

To explore elite bellicosity, I investigate behavior during 1964-1966, the leadup to the Vietnam War, and the breakdown of the "Cold War Consensus" \citep{krebs2015dominant}.  The era is particularly interesting because, while the United States increased its commitment to Vietnam, it also engaged in several other interventions around the world \citep[ch. 16]{herring2008colony}. Thus, we would expect to see that an increase in bellicosity in the FRUS corpus would be correlated with an increase in hostile actions by the United States.

\begin{figure}[t!]
\centering
\includegraphics[scale=0.6]{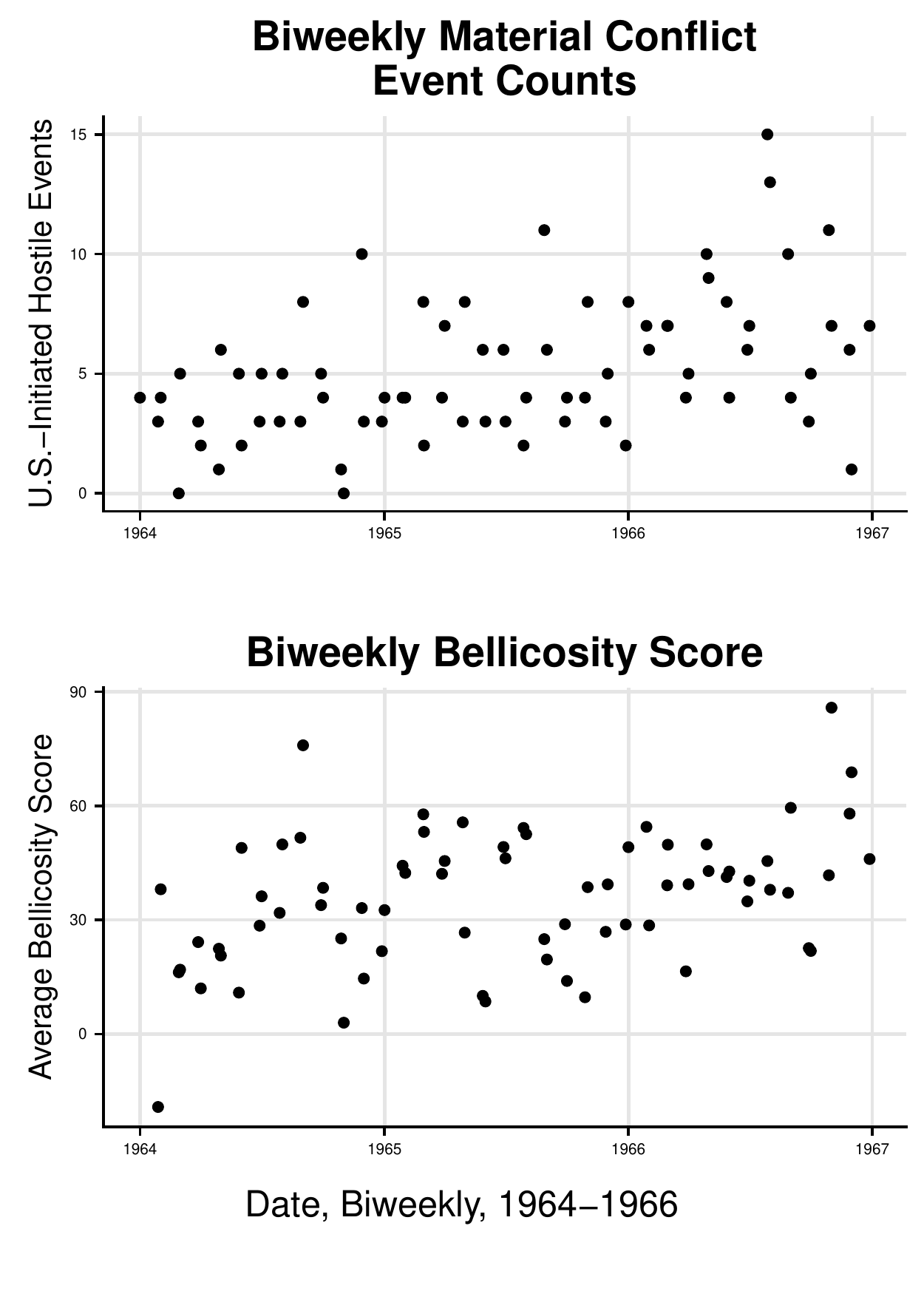}
\caption{Plotting material conflict event counts and bellicosity scores over time aggregated at the bi-weekly level. Both bellicosity and the count of material conflict events increase with time, as the United States increased its involvement in the Vietnam War. }
 \label{fig:descriptive_plot}
\end{figure}

I measure hostility using the Cline Center Historical Event Data, coded from the \textit{New York Times} \citep{cline2017event}. These data take the form \textit{(DATE, STATE A, ACTION, STATE B)}, where \textit{(STATE A, STATE B)} are directed dyads, \textit{DATE} is the day the event was observed, and \textit{ACTION} is one of five categories of action: neutral, verbal cooperation, verbal conflict, material cooperation, or material conflict \citep{norris2017petrarch2}. I select only those events where \textit{STATE A} is the United States, and sum events at the biweekly level. I measure hostility using counts of material conflict events, and display the hostile event counts in Figure \ref{fig:descriptive_plot}. 

To calculate bellicosity, I first estimate a Bayesian Word Embedding model, with context window of 9, $K=50$, keeping any word that occurs at least 40 times. I then anchor the embeddings on a ``war-peace" dimension. I summarize the results of the anchoring using Uniform Manifold Approximation and Projection for Dimension Reduction (UMAP), which calculates a low dimensional number of components, similar to principal components analysis. Unlike PCA, UMAP calculates distance using cosine similarity, while balancing both global and local structure in the embeddings \citep{becht2019dimensionality}. I present results in Table \ref{tab:umap_table}, and the components reveal themes in the corpus, clustering by region and issue, helping highlight the face validity of the embeddings.

\begin{table*}[t!]
\centering
\scalebox{0.85}{%
\begin{tabular}{llllllll}
  \hline \hline  
iran & doubtful & communications & tam & bases & relatively & robertson & initials \\ 
  iranian & blocked & relations & systematically & family & zambia & outflows & secretary \\ 
  shah & sponsored & appreciably & north & leave & udi & payments & footnotes \\ 
  aram & ultimatum & masses & hanoi & deployments & tran & liabilities & present \\ 
  iranians & telecommunications & sites & recce & fixed & rhodesia & fowler & conflict \\ 
  afghan & recommendation & overtures & drv & precondition & sr & banks & president \\ 
  squadron & imminent & concurs & vinh & laotian & neighboring & tax & raymond \\ 
  mnd & jet & harass & chau & reasons & continent & corporations & even \\ 
   
   \hline \hline
   \end{tabular}}
  \caption{The top words from a subset of components estimated from UMAP. Components include a variety of regional and substantive themes. These results help highlight the validity of the embeddings: semantically similar words are appearing near each other in cosine space.
  }
 \label{tab:umap_table}
\end{table*}

To estimate the bellicosity of a given document, I multiply the war-peace dimension by the document term matrix, averaging document bellicosity scores at the bi-weekly level. I plot the bi-weekly bellicosity scores in Figure \ref{fig:descriptive_plot}. 

\begin{figure}[t!]
\centering
\includegraphics[scale=0.4]{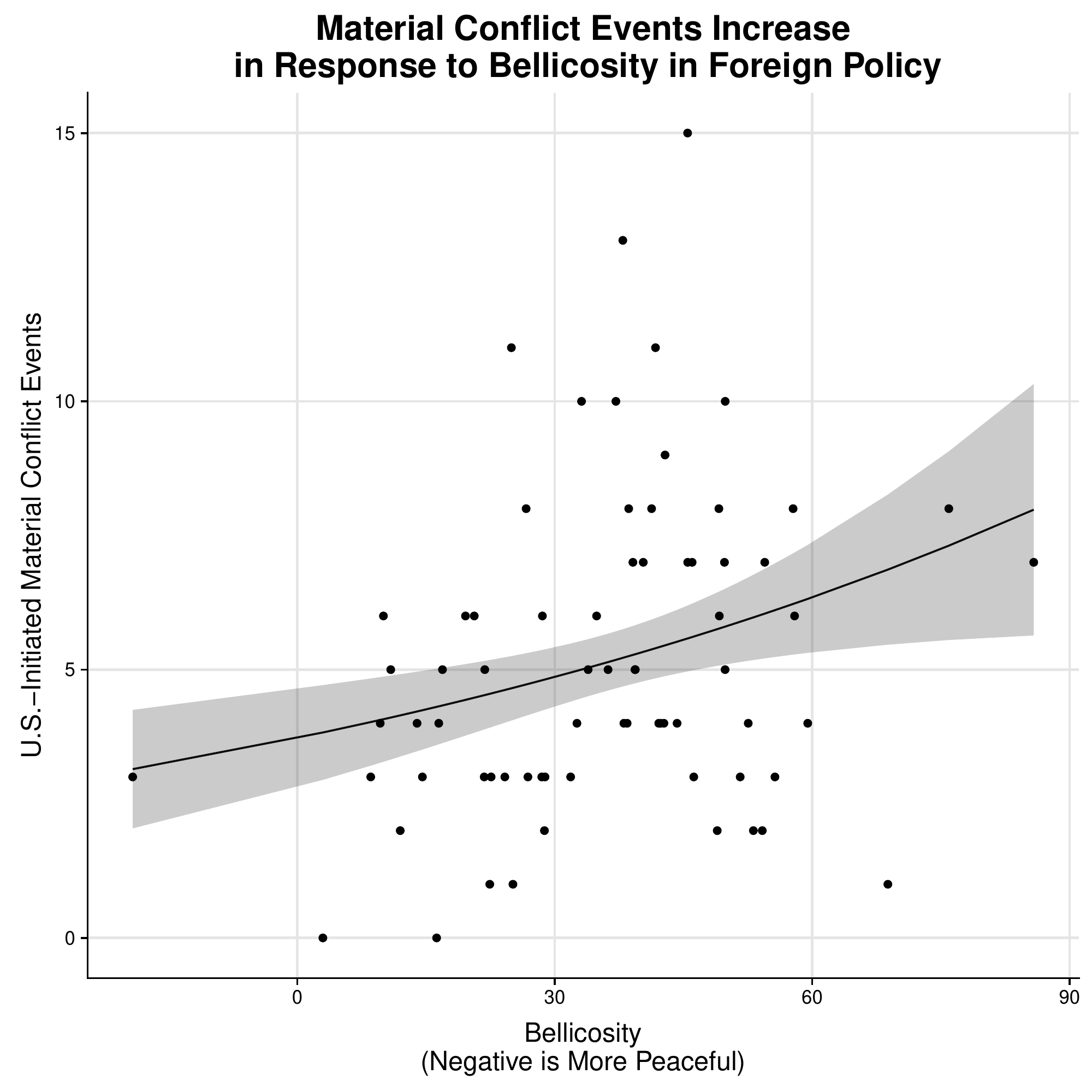}
\caption{An increase in the previous bi-weekly period's bellicosity is associated with an increase in U.S.-initiated hostile events. The regression is from a Poisson generalized linear model, and uncertainty is displayed with 95\% confidence intervals. }

 \label{fig:predicted_bellicosity_plot}
\end{figure}

To determine if there is a relationship between hostile events and bellicosity, I regress events on the lagged bellicosity (to account for a delay in policy implementation), using a Poisson generalized linear model, due to the count-distributed nature of the outcome.\footnote{In the appendix, I remove outliers and high-leverage points from the dataset, and fit the model again. Results do not change.} I plot the regression line against the data in Figure \ref{fig:predicted_bellicosity_plot}, and find a positive and statistically significant effect.

This result suggests that bellicosity in elite deliberations, captured in diplomatic documents, results in an increase in conflictual events, which suggests that the documents in the \textit{FRUS} corpus do not simply contain cheap talk, these deliberations ultimately shape policy. These findings also help establish the validity of the ``bellicosity'' scale, that is, it correlates with an entirely separate dataset, which captures a similar phenomena. All replication materials are available at \url{https://github.com/adamlauretig/bwe_application_naacl_2019}.

%outline:\\
%Introduce FRUS \\
%Describe period of study \\
%Anchor embeddings, descriptive cosine similarity \\
%Describe outcome: cline center events \\
%Create ``bellicosity scores'' \\
%Regress mids on bellicosity scores\\

\section{Conclusion}

In this paper, I introduced Bayesian Word Embeddings, a method for estimating word embeddings which uses variational bayesian methods. I incorporated  Automatic Relevance Determination priors on the embedding dimensions, relaxing the requirement that all dimensions have equal weight. Linking word embeddings to Bayesian latent variable models, I then discussed issues with identification, and solutions proposed in the ideal-point literature, as well as offering an alternative which allows for scaling along dimensions of interest, which creates model model that can then be used in a regression.

I applied Bayesian Word Embeddings to two cases: examining the change in American attitudes about the world before and after 1945 as captured in Presidential inaugural addresses, and then, testing whether an increase in the bellicosity of internal elite discussion (in diplomatic documents) results in an increase in American hostility. I found that there was a statistically significant different in the views of the world expressed in inaugural addresses, and that this shift was the opposite of what hypotheses generated from international relations theory would expect. When testing the effect of bellicosity on the hostility of American foreign policy, I that an increase in bellicosity resulted in an increase in hostility. 

Overall, I have contributed a tool which can serve many purposes for social scientists. By building a probabilistic embedding model, I have constructed a tool which moves beyond document-based inferential approaches to text as data, allowing inference on individual words. This promises new reaches for social scientists, in particular, the promise of crossover with interpretivist work, building on \citet{nelson2017computational}. Concepts such as securitization theory \citep{waever1995securitization} draw on the idea that language and word choice by elites shape the attitude of the public, and through the methods introduced above provide the opportunity to generate statistical tests for hypotheses derived from theories like securitization theory.

Future methodological work will follow three tracks. The first will build on \citet{rudolph2016exponential} and \citet{han2018conditional}, one goal is incorporating document-level metadata into embedding estimation, allowing embeddings to vary according to document-specific attributes, and then, identifying the resulting embeddings. The second will take advantage of stochastic variational inference \citep{hoffman2013stochastic} to enable Bayesian Word Embeddings to scale to massive corpora. Finally, the third track for future word will involve tying the anchoring approach discussed above with the emerging literature on making casual claims from text \citep{fong2016discovery, mozer2018matching}, and taking advantage of the word similarities to identify appropriate linguistic counterfactuals. 

\section*{Acknowledgments}
I would like to thank Bear Braumoeller, William Minozzi, and Gregory Smith, along with the three reviewers for valuable feedback on this project. I would like to thank Michael Neblo and the Institute for Democratic Engagement and Accountability at Ohio State for funding which helped support this project. \\

%\clearpage

\bibliography{embeddings}
\bibliographystyle{acl_natbib}

\clearpage

\appendix

\section{Comparing BWE to STM}
\label{sec:appendix}
Unlike the results from the Bayesian Word Embedding, the structural topic model detects no difference in topics before and after 1945. The top words, as determined by FREX score, are visible in \ref{tab:stm_inaug_tab}.

\begin{figure}[t!]
\centering
\includegraphics[scale=0.4]{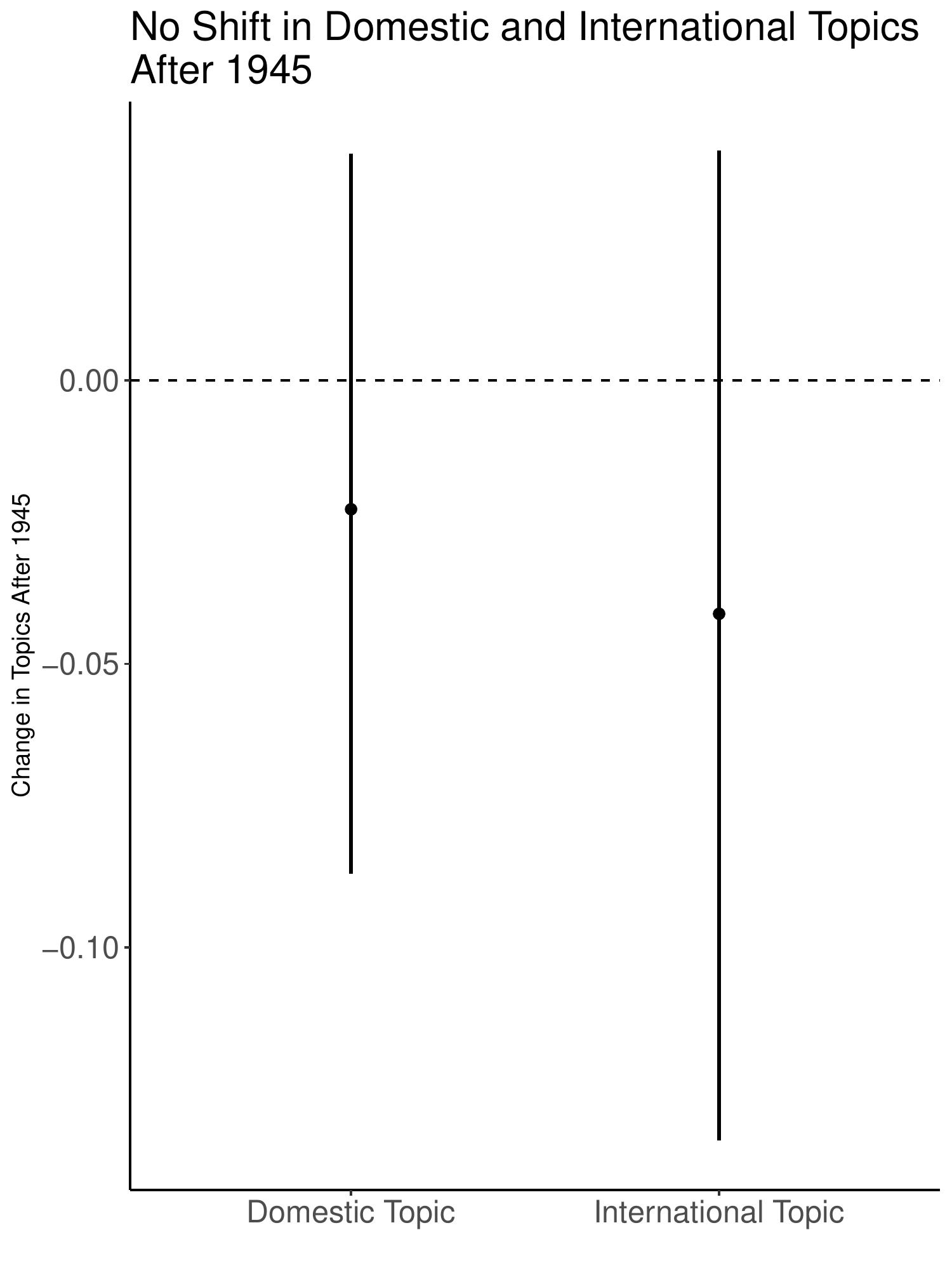}
\caption{There is no significant difference between foreign and international topics before and after 1945, uncertainty is displayed with 95\% confidence intervals. }

 \label{fig:stm_inaug}
\end{figure}

\begin{table}[ht]
\centering
\begin{tabular}{ll}
  \hline
International Topic & Domestic Topic \\ 
  \hline
representative & pacific \\ 
  civilization & territory \\ 
  making & question \\ 
  international & whilst \\ 
  tax & importance \\ 
  popular & constitution \\ 
  concern & slavery \\ 
  supreme & domestic \\ 
   \hline
\end{tabular}
\caption{Top eight words from structural topic model for international and domestic topics, by FREX score.}
\label{tab:stm_inaug_tab}
\end{table}

\section{Inauguration Robustness Check}

\begin{figure}[t!]
\centering
\includegraphics[scale=0.4]{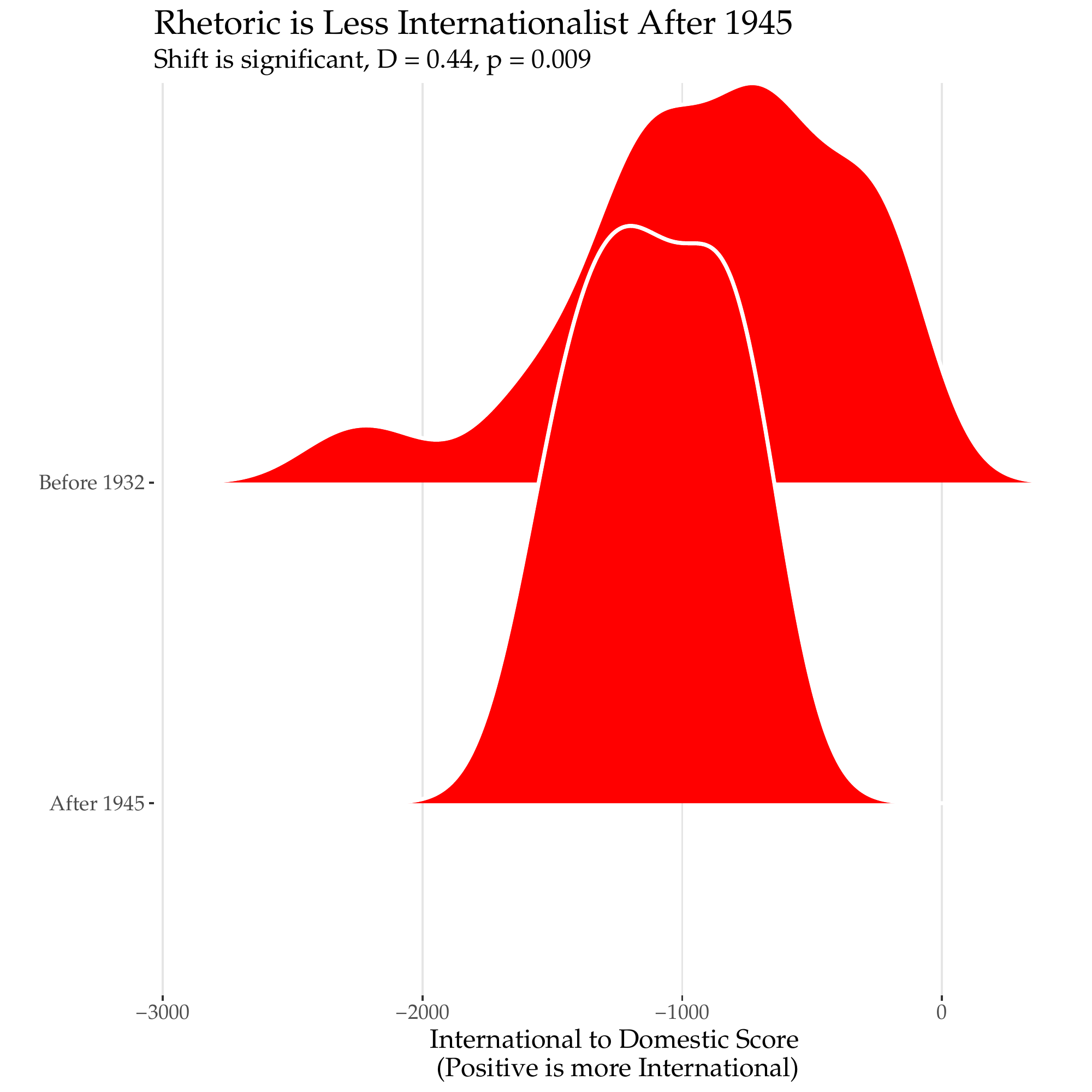}
\caption{Even excluding the Roosevelt administration, when only examining inaugural addresses from before 1932 and after 1945, the pre-1932 inaugural addresses are more internationalist than the post-1945 addresses.}

 \label{fig:no_fdr}
\end{figure}

One concern with the role of internationalism in inaugural addresses is that by splitting at 1945, the ``internationalism'' of the pre-1945 sample is due to World War Two, and the Roosevelt presidency. To account for this, I re-estimate the Kolmogorov-Smirnov test from above, excluding the Roosevelt inaugural addresses, and present the results in \ref{fig:no_fdr}. This compares inaugurals from before 1932 to those after 1945, and the substantive results do not change.

\section{GLM without Outliers}

\begin{figure}[t!]
\centering
\includegraphics[scale=0.4]{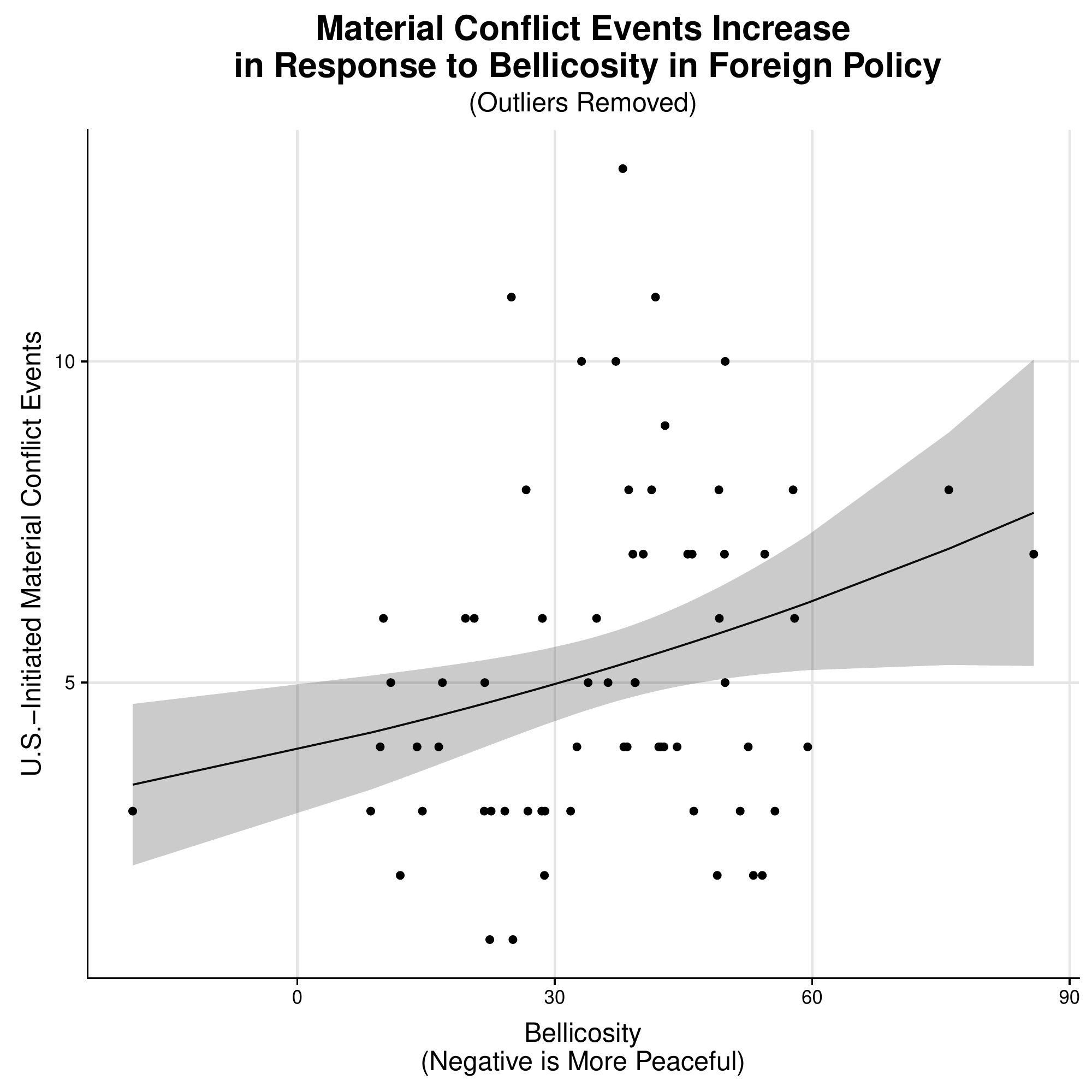}
\caption{An increase in bellicosity is associated with an increase in U.S.-initiated hostile events. The regression is from a Poisson generalized linear model, and uncertainty is displayed with 95\% confidence intervals. }

 \label{fig:predicted_bellicosity_outliers}
\end{figure}

To ensure that the results presented in Figure \ref{fig:predicted_bellicosity_plot} were not the results of outliers, I removed any outliers and high-leverage points, and re-fit the model. The results were the same, as visible in Figure \ref{fig:predicted_bellicosity_outliers}.

\end{document}